# HANDS: a multimodal dataset for modeling toward human grasp intent inference in prosthetic hands

Mo Han[1] · Sezen Yağmur Günay[1] · Gunar Schirner[1] · Taşkın Padır[1] · Deniz Erdoğmuş[1]

**Abstract**

Upper limb and hand functionality is critical to many activities of daily living, and the amputation of one can lead to significant functionality loss for individuals. From this perspective, advanced prosthetic hands of the future are anticipated to benefit from improved shared control between a robotic hand and its human user, but more importantly from the improved capability to infer human intent from multimodal sensor data to provide the robotic hand perception abilities regarding the operational context. Such multimodal sensor data may include various environment sensors including vision, as well as human physiology and behavior sensors including electromyography and inertial measurement units. A fusion methodology for environmental state and human intent estimation can combine these sources of evidence in order to help prosthetic hand motion planning and control. In this paper, we present a dataset of this type that was gathered with the anticipation of cameras being built into prosthetic hands, and computer vision methods will need to assess this hand-view visual evidence in order to estimate human intent. Specifically, paired images from human eye-view and hand-view of various objects placed at different orientations have been captured at the initial state of grasping trials, followed by paired video, EMG and IMU from the arm of the human during a grasp, lift, put-down, and retract style trial structure. For each trial, based on eye-view images of the scene showing the hand and object on a table, multiple humans were asked to sort in decreasing order of preference, five grasp types appropriate for the object in its given configuration relative to the hand. The potential utility of paired eye-view and hand-view images was illustrated by training a convolutional neural network to process hand-view images in order to predict eye-view labels assigned by humans.

✉ Mo Han
han@ece.neu.edu

Sezen Yağmur Günay
gunay@ece.neu.edu

Gunar Schirner
schirner@ece.neu.edu

Taşkın Padır
tpadir@ece.neu.edu

Deniz Erdoğmuş
erdogmus@ece.neu.edu

[1] Northeastern University, 360 Huntington Ave, Boston, MA 02115, USA

## 1 Introduction

To enhance the quality of life of individuals with upper limb loss, further research to improve robotic prostheses is necessary. Along this direction, some electromyography (EMG)-based human intent inference solutions demonstrated promising results for patients with hand and wrist amputations [2,9,10]. The quality of EMG signals may dramatically vary across individuals depending on the specifics of their amputation. Some studies also investigated the possibility of complementing EMG with electroencephalography (EEG) signals to improve inference accuracy [2,9,15,19]; however, in many practical situations the added information from EEG does not have significant impact on performance. Both EMG and EEG models may need frequent calibration to account for signal nonstationarity due to various factors, such as electrode locations or skin conductance. Furthermore, due to the differences of amputations across individuals, the generalization of models for different users has been challenging. The introduction of visual evidence to the process may help alleviate some of these issues associated with physiological signals.

Differently from robot grippers studies [16,20] which are primarily concerned about identifying gripper approach directions and gripable handles, dexterous prosthetic hand research [7,8] needs to address interactive and shared control aspects associated with the collaborative human–robot grasping. A prosthetic hand does not have the freedom to choose its approach trajectory decided by the individual; therefore, it will be primarily concerned with the inference and execution of the human's intended grasp, as estimated from available measurements of the environment and physiology. With the advances made in object recognition from images using convolutional neural network (CNN) models [4,14,25,26], prosthetic hand researchers started exploring the use of CNNs as a visual evidence processing solution [6–8,27]. Most such studies mainly focus on arriving at a grasp type classification label based on an image of the object of interest from the eye-view of the human. These approaches benefit from the availability of pretrained CNN models which could be widely applicable to different scenes since the visual information could be easily obtained and the trained models are not user-based, and thus improve the generalization across individuals.

Large datasets are necessary to train deep CNNs. Databases for image classification, such as ImageNet [4], are generally aimed for visual object recognition, therefore they contain object categories as labels. For grasp type classification from images of objects, the labels need to be appropriate grasp types. However, there are few datasets available for the grasp classification task for a five-finger prosthetic hand. Ghazaei et al. [6] manually defined a mapping from object categories to grasp types based on the COIL100 dataset. They used a similar method in a follow-up study [7]. Even though the dataset contains images of objects with irregularly shapes, their mapping that assigns one grasp type to each object category is independent of the orientation when approaching the complex-shaped object. Since appropriate grasp type may depend on the relative orientation of the object to the hand, it is possible that this dataset is not appropriate to develop image analysis models for prosthetic hand applications. Their datasets also do not appear to be openly available to others (Fig. 1).

In this paper, we present an open dataset that we acquired, got labeled by multiple humans, and curated with the intention to train a CNN for grasp type estimation from images of various objects with complex shapes observed in different orientations. This *HANDS* dataset[1] contains both eye-view and hand-(palm-)view images of objects along with grasp types associated with them according to multiple human

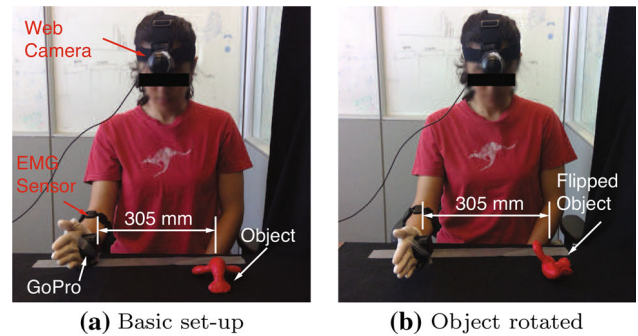

**(a)** Basic set-up                    **(b)** Object rotated

**Fig. 1** Data collection: (1) a webcam worn on a headband and a GoPro attached to the palm provide human eye-view and hand-view videos; (2) EMG was acquired with a MYO armband. The hand started 12 inches (305 mm) to the right side of the object, and a ruler with 0.5 inch (12.7 mm) ticks was positioned on the table as seen in the pictures. Between trials, the object was horizontally and vertically rotated to capture views for multiple orientations

labelers observing the scene from the eye-view images. The intention is to train robot-vision models, such as CNNs that can operate on hand-view images of an object to estimate the grasp type label assigned by humans who see the same object from their eye-views. To validate that this is achievable, we trained a CNN with hand-view images as the input, and eye-view-based labels as the desired output. Moreover, for the multimodal control study, the real-time videos from eye-view and hand-view along with the corresponding EMG signals and forearm acceleration, angular velocity were collected with an inertial measurement unit (IMU) during the hand motion.

## 2 Methodology

Recent studies that acquired visual data in the context of prosthetic hand research mostly used eye-level cameras worn by the user/subject, typically located on the head [6,7], possibly because they used off-the-shelf robot hands that did not have cameras built-in. While it is possible that future prosthetic hands might have paired head/body-worn cameras that provide visual evidence for human intent inference and environment state estimation, we contemplate prosthetic hands that have cameras built-in to the robot itself for the system to be self-contained. Arguably, cameras worn on the user's head that provide an eye-view of the scene, and cameras located on the prosthetic hand that provide a hand-view of the scene can both have useful information to facilitate state estimation and intent inference; likewise, they can introduce different challenges, such as in cases where the user does not look at the object while moving the hand to grasp it.

DeGol et al. [3] presented a prosthetic hand into the palm of which a camera is embedded. They trained a CNN to classify grasp types for objects as seen from an eye-view image,

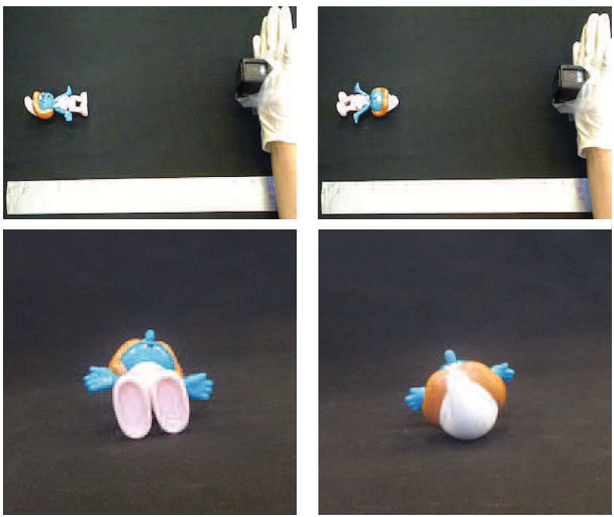

**Fig. 2** Images of both eye and hand perspectives from different orientations: we rotated the object to different orientations to catch the shape information from various approaching directions

but then used this CNN to infer intended grasp types based on the hand-view images produced by the palm-camera in the prosthetic hand. We assert that a grasp type inference model that operates on hand-view images will perform better if it is trained using data that show objects from the hand-view. Moreover, the user forms the intended grasp type based on their eye-view observation of the hand-object scene. Consequently, an image dataset where human labelers determine appropriate grasp types for objects as seen from the eye-view, paired with matching images of the object/scene captured from a camera on the prosthetic hand-view, is potentially very useful to develop vision-based human intent estimators for prosthetic hands that contain built-in cameras.

The role of the prosthetic hand is to serve its user by executing grasps as intended and appropriately. A prosthetic hand, with its built-in cameras seeing the object from the hand-view, needs to become informed about the human intuition regarding manipulation of objects within the user's personal space, as well as the user's experience-based ability to take into account the irregular shapes and orientations of objects in space. We collected, and present here, a dataset that includes pairs of images from eye-view and hand-view, of many objects, at different orientations relative to the hand. Figure 2 illustrates this with one object in two orientations. The image pairs were labeled by multiple humans observing only the eye-view scene, while the prosthetic hand which will only have access to the hand-view will rely on an image classifier that can be trained to predict the intended grasps from the hand-view images. This proposition is different from what DeGol et al. [3] did, even though they had access to a hand with a camera in its palm.

## 2.1 Set-up

The set-up for data collection is shown in Fig. 1. A subject was equipped with a headband that held a camera to capture the human eye-view images, and a glove that held a camera to capture the (right-)hand-view images. An object is placed on the table with some arbitrary orientation. With the help of a ruler placed on the table, the subject positioned the hand approximately 12 inches (305 mm) away from the object, with the palm facing it. At this initial position, simultaneous pictures from the eye-view and hand-view cameras were captured. Following an audio cue, the subject started moving the hand toward the object and grasped it as appropriate for the particular object and its orientation, lifted the object from the table briefly, put the object down, and retracted the hand back to the initial pose and position. During this motion, (approximately) synchronous videos from both eye-view and hand-view were captured, along with EMG and IMU. Paired eye-view and hand-view images of four sample objects are shown in Fig. 3.

The initial positions and spatial separation between the object and the hand were selected such that (1) they both were visible in the eye-view image throughout the process, and (2) the trial lasted long enough to capture approximately 3 seconds of EMG/IMU data during motion. The procedure was repeated a number of times, each time rotating the object vertically and horizontally, as can be seen in Fig. 1, to capture several possible relative orientations of the object with respect to the hand. The number of orientations varies depending on the shape of each object.

A black background and a white glove were used to simplify the problem of identifying the hand in images and videos. The hand-view images and videos were captured with a GoPro Hero Session camera (3648×2736 pixels). The eye-view images and videos were captured with a Logitech Webcam C600 camera (1600×1200 pixels). The ruler placed on the table for reference had ticks at every half inch. The EMG and IMU measurements from the arm were acquired with a MYO armband (Thalmic Labs Inc.) that was placed on the right upper arm of the subject.

The dataset consists of 413 such instances of multimodal data acquired using 102 different ordinary objects, such as office supplies, utensils, stuffed animals, and other toys.

## 2.2 Data annotation

Unlike object classification problems in computer vision, appropriate grasp type in a given setting may depend on personal preferences of humans as well as other factors. Consequently, for a given eye-view image, there may not be a single correct grasp label. In order to capture this interperson variability, 11 individuals were asked to sort five hand grasp types in decreasing order of preference to grab the object as seen in each eye-view image. The label acquisition process was carried out using an interface developed with Visual Studio 2013, which randomly ordered the images and allowed them to indicate their order of preference for grasp types.

**Fig. 3**  **a–d** Eye-view images taken by Webcam; **e–h** corresponding hand-view images taken by GoPro

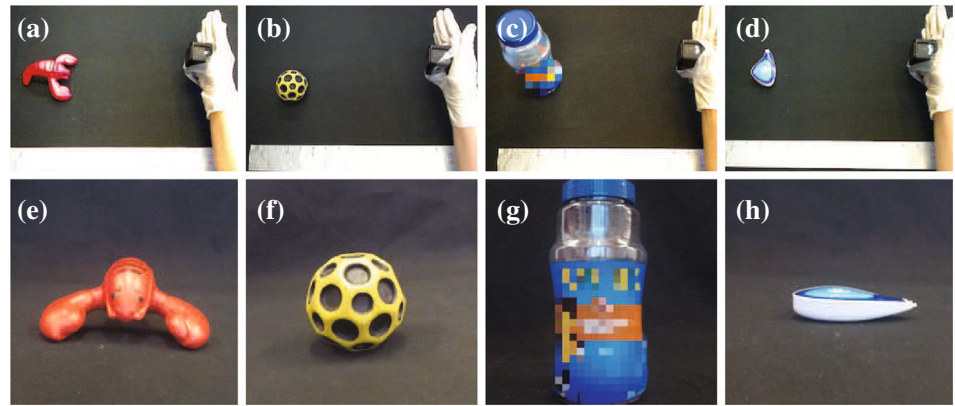

Feix et al. [5] proposed that human grasp taxonomy consists of 33 classes if only the static and stable grasp types are taken into account. The human hand has at least 27 degrees of freedom (DoF) to achieve such a wide range of grasp types; however, most existing prosthetic hands do not have this many DoF. For instance, thumb abduction versus adduction, which is a crucial capability of the human hand, is a challenging robotic design problem, and some robotic hands resort to unpractical approaches like using a manually controlled switch to alternate between these movements [1]. Even contemporary state-of-the-art prosthetic hand designs [22] have much fewer DoF than the human hand, which limits their number of feasible grasp types to 14.

Gunay et al. [9] showed that with appropriate multichannel EMG signals, these grasp types that are feasible for state-of-the-art prosthetic hands can be classified accurately. While EEG alone may not be capable of discriminating a large number of intended grasp types, supplementing EMG evidence with EEG may lead to improved intent inference in the case of some amputations. Ozdenizci et al. [19] demonstrated that five particular hand gestures (four of them grasp types) can be reliably inferred from EEG, and they provide a reasonable initial coverage of the grasp taxonomy variety established by Feix et al. [5]. Specifically, these five grasp types, shown in Fig. 4, are: open palm, medium wrap, power sphere, parallel extension, and palmar pinch. This list includes three thumb-abducted, 1 thumb-adducted grasp types, as well as a typical resting state of the hand.

### 2.3 Data augmentation and pre-processing

Data augmentation methods have been successfully used in machine learning, especially computer vision, to promote invariance in models for certain anticipated variations in input data, such as illumination, sharpness, background clutter, and position of object of interest in the image. In addition to the original 413 hand-view images, the dataset also provides synthetically manipulated versions of these. A total of 4466 synthetic hand-view images were generated by first segmenting the object, then applying a random blurring filter (a square-shaped moving average filter with a randomly selected impulse response length) to the object segment, then randomly translating the object segment in the image, and finally adding to the background an approximately Gaussian distributed noise with zero mean and a random variance for each of the red, green, and blue color coordinates. Some sample synthetic images are shown in Fig. 5. Each synthetic image was assigned one of the human labelers' top choice for that instance, without replacement (each label was used at most once for augmented images of an instance). For each synthetic image, the bounding box of the object location was also included in the dataset.

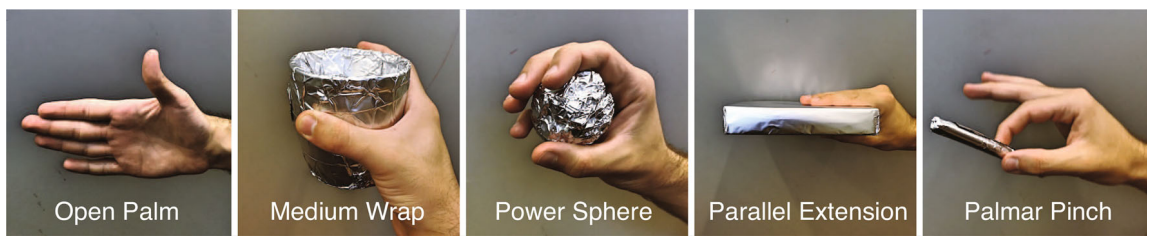

**Fig. 4**  Five grasp types selected for the classification problem, sorted by labelers in decreasing order of preference

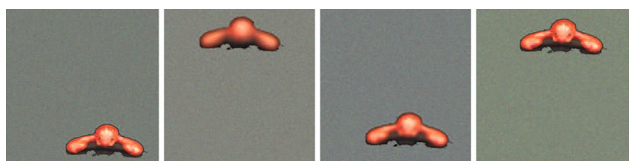

**Fig. 5** Data augmentation: adding random background noise, and randomly blurring and locating the object

## 2.4 Gesture classification

In order to illustrate the utility of having paired eye-view and hand-view images, where the appropriate grasp types for a given object/hand initial state were assessed by multiple humans, a CNN classifier was trained to predict the top choice of a human labeler from the hand-view image of the object. CNNs have been very successful in image classification problems [14], and pretrained CNNs can be used to initialize the grasp type classifier, which will be fine-tuned with the available labeled data. CNNs are interpreted as extracting visual features from the input image through their early convolution, pooling, and normalization layers [25]. Specifically, VGG-16 that consists of 16 convolutional layers and about 138 million parameters [26] will form the basis of our CNN grasp type classifier, because it has favorable performance when compared to some other deep neural network models [17,21]. A version of VGG-16 that has been trained on ImageNet [4] object classification dataset, with 1000 object categories, can be found online [8].

The last layer of VGG-16 was replaced by a fully connected layer with sigmoid nonlinearities and five outputs, and of the resulting CNN, the last three layers were trained using hand-view images of randomly selecting 80% of the unique objects to predict their associated grasp type labels. The images of the 20% of the unique objects were used for validation. The training procedure used stochastic gradient descent for binary cross-entropy [23] as the loss function. Furthermore, to cope with the overfitting, we added zero padding layers before each convolutional block, and L2 regularization terms were included in the objective function for the weights of every convolutional layer. The process is illustrated in Fig. 6.

## 2.5 Evaluation

The performance of the CNN was evaluated in terms of both loss function value and grasp type classification accuracy. The loss and accuracy values on both training and validation data sets are shown in Fig. 7. It can be observed that both of these measures show rapid improvement in early epochs, and progress slows down significantly before epoch 30. Both loss and accuracy values on training and validation sets are similar, which may indicate the prevention of overfitting. The accuracy of the trained CNN was approximately 89%, and its loss value was approximately 0.04. These results demonstrate that the hand-view images are informative about human grasp type intent that is based on eye-view images of the object.

The accuracy achieved for power sphere and medium wrap, 91.2% and 90.5%, respectively, was higher compared to the other grasp types. One possible reason for this could be that more objects are labeled as power sphere or medium wrap (large class priors). Another reason could be the single 80% versus 20% training, and validation dataset split was favorable to these class labels. In future work, the dataset could be extended to include objects that lead to a more balanced grasp type class distribution.

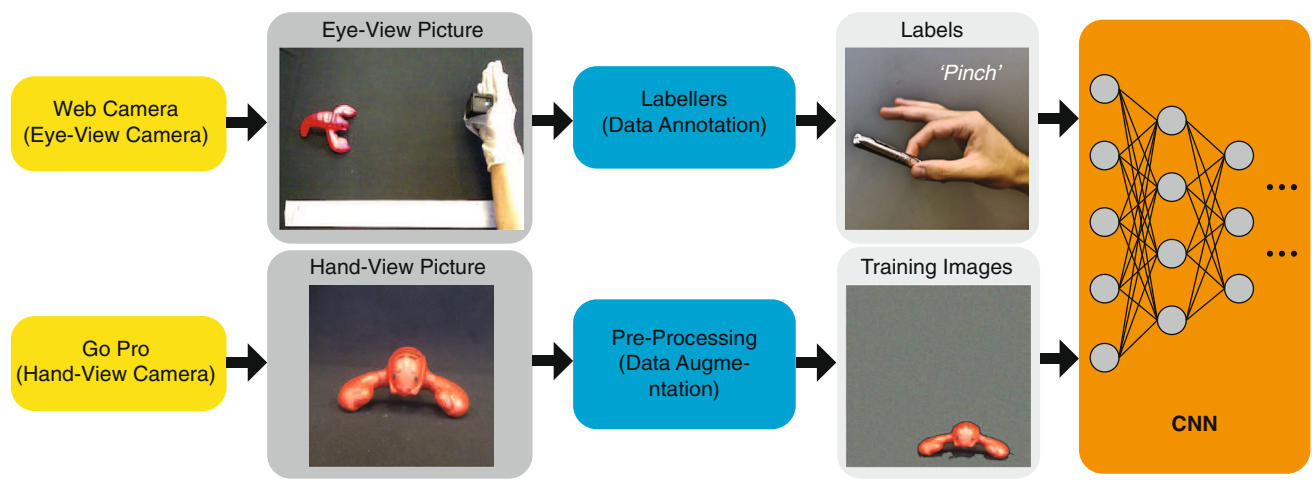

**Fig. 6** An illustration of the data acquisition, labeling, and classifier training process: (left) images from eye and hand perspectives acquired simultaneously; (2) multiple human labelers assigned grasp type labels to each image, based on the eye-view images; (3) after data augmentation, a convolutional neural network was trained to predict grasp type label from the hand-view images

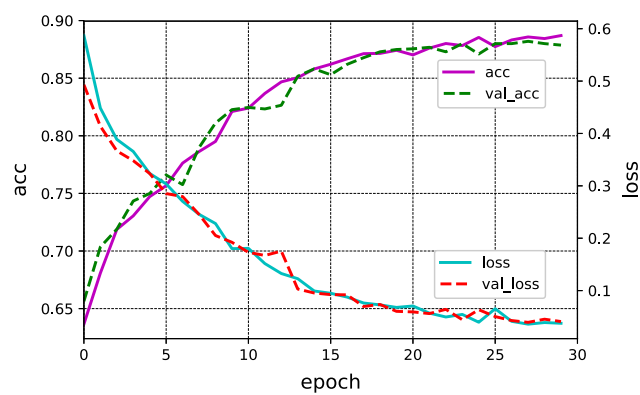

**Fig. 7** Accuracy and loss curves of the CNN during training on training and validation datasets. The acc and val-acc are the accuracies of training and validation data; the loss and val-loss are their loss values

# 3 Further usage of the dataset

The HANDS dataset also includes the video of various objects from both hand and eye-views. Additionally, EMG and IMU signals from the forearm are available for each trial. This additional data can support further research in multimodal grasp type intent inference.

## 3.1 The usage of the visual information

An alternative use of eye-view and hand-view paired images is the use of the CNN to provide a probability distribution for all possible grasp types, based on the hand-view image [11]. Such a CNN could be further extended to process the videos during the approach to the object, in order to generate a sequence of grasp type probability distributions over time. This probability distribution sequence can be fused with EMG and other evidence to obtain a grasp type posterior probability distribution that evolves over time along the hand trajectory. Depth evidence can be extracted from the video using computer vision techniques [13,24], in order to help inform the prosthetic hand motion planner to facilitate appropriate timing of grasp execution, to simulate the potential benefit of having a depth camera built-in to the hand.

## 3.2 The usage of the bounding box

The bounding box of the object in each image can help identify the object for different augmentation procedures. Different CNNs, such as YOLO [21] and SSD [17], can be trained using these images for lighter solutions that facilitate embedded real-time implementations. Additional object coordinate estimators could be trained to help with motion planning and controller actuation in prosthetic hands.

## 3.3 The usage of the EMG data

Even though visual evidence from hand or eye-views can provide strong priori expectations about what grasp type the user of a prosthetic hand might execute, ultimately, the intended grasp type of the human cannot be extracted solely based on the shape and orientation of the object in the view. The EMG data from the forearm will provide strong evidence regarding human grasp intent, especially as the hand gets closer to the object, and the muscles contract in anticipation of grasp execution. For instance, power sphere could be the most likely grasp type for a ball according to visual evidence, but the user might have the intent to push the ball rather than grasp it. Access to EMG is essential to distinguish between what grasp is likely a priori based on shape, and what the human intends to do in a specific instance. This dataset enables EMG-video fusion such as what Gigli et al. presented [8].

## 3.4 The usage of the IMU data

Prosthetic hands do not have control over the approach trajectory to the object; therefore, having access to independent kinematic measurements may be useful for motion planning and control when solving the shared control problem from a robotics perspective. The dataset contains three-dimensional orientation, acceleration, and angular velocity extracted from IMU measurements. Human reach-to-grasp trajectories exhibit tightly coupled behavior between hand velocity and aperture opening and access to IMU data can help use such dependencies in human reach-to-grasp trajectories [12,18].

# 4 Conclusion and future work

This paper presents an open multimodal dataset for human grasp intent inference using visual, physiological, and kinematic measurements. A unique aspect of the dataset is the presence of paired eye-view and hand-view images of an object, with multiple human labeler preferences for grasp types sorted by order of preference based on eye-view images. The utility of such image pairs was illustrated by training a CNN classifier that predicts eye-view-based grasp type labels from hand-view images. The dataset also includes videos and forearm EMG and IMU signals during the trials that consist of approach, grasp, lift, put-down, and retract phases. The dataset is available online. A future enhanced version of this dataset will include data for more unique objects, with more cameras on the prosthetic hand including depth sensing, and better EMG with anatomically informed electrode placement. Intent inference will rely on augmenting prior estimates based on streaming visual evidence with EMG and other signals.

**Acknowledgements** This work was supported by NSF (CPS-1544895, CPS-1544636, CPS-1544815), NIH (R01DC009834).